\documentclass[10pt,twocolumn,letterpaper]{article}

\usepackage{cvpr}              
\usepackage{url}
\usepackage{natbib}
\usepackage{amsmath}
\usepackage{caption}
\usepackage{booktabs}      
\usepackage{multirow}
\usepackage{graphicx}
\usepackage{float}
\usepackage{pifont}
\usepackage{url}            
\usepackage{booktabs}       
\usepackage{amsfonts}       
\usepackage{nicefrac}       
\usepackage{microtype}      
\usepackage{amsmath, amssymb, amsthm}
\usepackage{colortbl}
\usepackage{pifont} 
\usepackage{multirow}
\usepackage{multicol}
\usepackage{makecell}
\usepackage{graphicx}
\usepackage{diagbox}
\usepackage{subcaption}
\usepackage[utf8]{inputenc}
\usepackage{svg}
\usepackage{bm}
\usepackage{comment}

\usepackage{float}
\usepackage{enumitem}
\usepackage{caption}
\usepackage{graphicx}
\usepackage{caption}
\usepackage{booktabs}
\usepackage{graphicx}    
\usepackage{caption}     
\usepackage{subcaption}  
\usepackage[ruled,vlined]{algorithm2e}
\definecolor{cvprblue}{rgb}{0.21,0.49,0.74}
\usepackage[pagebackref,breaklinks,colorlinks,allcolors=cvprblue]{hyperref}

\def\paperID{} 
\def\confName{CVPR}
\def\confYear{2026}

\title{Forecast the Principal, Stabilize the Residual: Subspace-Aware Feature Caching for Efficient Diffusion Transformers}

\author{
\textbf{Guantao Chen}$^{1,2}$\quad
\textbf{Shikang Zheng}$^{1,3}$\quad
\textbf{Yuqi Lin}$^{1,4}$\quad
\textbf{Linfeng Zhang}$^{1,\dag}$\\
\textsuperscript{1}Shanghai Jiao Tong University \quad 
\textsuperscript{2}Sun Yat-Sen University  \\
\textsuperscript{3}South China University of Technology \quad
\textsuperscript{4}Jilin University
}

\begin{document}
\maketitle
\renewcommand{\thefootnote}{\fnsymbol{footnote}}
\footnotetext[2]{Corresponding author.}
\renewcommand{\thefootnote}{\arabic{footnote}}
\begin{abstract}
Diffusion Transformer (DiT) models have achieved unprecedented quality in image and video generation, yet their iterative sampling process remains computationally prohibitive. To accelerate inference, feature caching methods have emerged by reusing intermediate representations across timesteps. However, existing caching approaches treat all feature components uniformly. We reveal that DiT feature spaces contain distinct principal and residual subspaces with divergent temporal behavior: the principal subspace evolves smoothly and predictably, while the residual subspace exhibits volatile, low-energy oscillations that resist accurate prediction. Building on this insight, we propose SVD-Cache, a subspace-aware caching framework that decomposes diffusion features via Singular Value Decomposition (SVD), applies exponential moving average (EMA) prediction to the dominant low-rank components, and directly reuses the residual subspace. Extensive experiments demonstrate that SVD-Cache achieves near-lossless across diverse models and methods, including 5.55$\times$ speedup on FLUX and HunyuanVideo, and compatibility with model acceleration techniques including distillation, quantization and sparse attention. Our code is in supplementary material and will be released on Github.
\end{abstract}
\section{Introduction}

Diffusion Transformer (DiT) models have demonstrated remarkable performance in both image and video generation, exhibiting powerful modeling capabilities and high-quality outputs, but their iterative denoising process incurs high computational cost. This substantial computational overhead poses challenges for deployment in latency-sensitive or resource-constrained environments, motivating ongoing efforts to develop more efficient inference techniques.

Two main acceleration strategies have emerged to address this: reducing the number of sampling steps through algorithmic innovations~\citep{lu2022dpm}, and decreasing per-step computational cost via architectural improvements~\citep{yuan2024ditfastattnattentioncompressiondiffusion, zhao2025realtimevideogenerationpyramid}. Among these approaches, training-free feature caching stands out. By leveraging temporal coherence in hidden representations and reusing cached features, it significantly reduces computational redundancy. Pioneering work like DeepCache~\citep{ma2024deepcache} validated this concept on U-Net models, subsequent methods including FORA~\citep{selvaraju2024fora}, ToCa~\citep{zou2024accelerating}, and TaylorSeer~\citep{liu2025reusingforecastingacceleratingdiffusion} extended caching to transformer architectures and formulated feature caching as temporal evolution of hidden states. These approaches, however, implicitly assume that the entire feature space is smooth and predictable, thus applying a single prediction strategy can model the whole space. Yet, the feature space of diffusion transformers is extremely high-dimensional and expecting a globally smooth temporal trajectory across all dimensions may be untenable. This conjecture motivates us to dive into the potential subspaces of diffusion features and treat each subspace accordingly, rather than enforcing a one-size-fits-all predictor over the whole space.

\begin{figure}[t]
  \centering
  \includegraphics[trim=45 170 315 185 clip,width=1\linewidth]{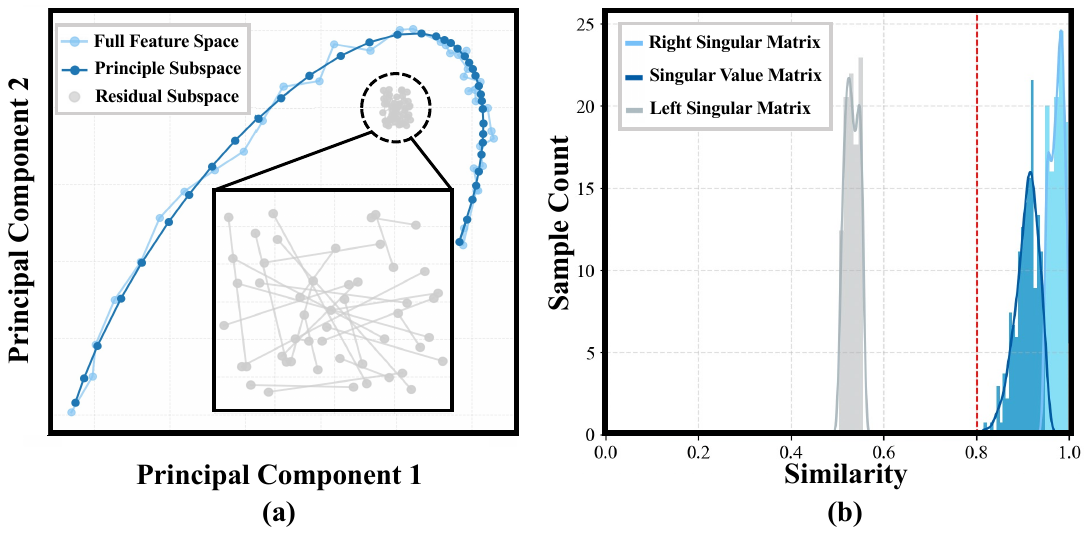}
  \caption{(a) PCA visualization of different feature spaces: Original Full Feature Subspace shows oscillatory trajectory while Principal Subspace evolves smoothly. Residual subspace is oscillatory and low-energy. (b) Singular values and right singular vectors are stable across different prompts. Similarity is evaluated using the product of cosine similarity and magnitude similarity. A similarity over 0.8 typically indicates a stable and reusable subspace.}
  \label{fig:intro}
\end{figure}

We begin by visualizing the trajectory of the full feature space. As shown in Fig.~\ref{fig:intro}(a), the trajectory is coherent yet noticeably oscillatory, confirming our intuition that global prediction is unreliable. We therefore decompose the feature space using singular value decomposition (SVD) into a rank-$k$ principal subspace and an orthogonal residual subspace. The principal subspace, as illustrated in Fig.~\ref{fig:intro}(a), evolves smoothly and predictably, thus supporting stable extrapolation. In contrast, the residual subspace displays high-frequency, low-energy fluctuations that are intrinsically hard to forecast and prone to error amplification. These observations argue for a nuanced, subspace-aware strategy: apply prediction where dynamics are stable (principal subspace) and simply reuse features where prediction is inherently unreliable (residual subspace).

Therefore, we propose \textbf{SVD-Cache}, a subspace-aware feature caching framework that decomposes DiT features via Singular Value Decomposition (SVD), applying distinct methods to different subspaces. SVD-Cache begins by decomposing the feature space to dominant and residual subspaces, and then applies EMA prediction to the dominant subspace while reusing the residual subspace directly. Intuitively, the decomposition needs to be performed across diverse prompts. However, we surprisingly uncover an intriguing property that for different prompts, the SVD decomposition yields \textbf{highly stable singular values and right singular matrices} as shown in Fig.~\ref{fig:intro}(b), though the left singular matrices do not show such invariance, they can be reconstructed from the current features at low cost. This invariance allows us to decompose the feature space on a reference prompt offline and cost-effectively reconstruct the decomposition for any input prompt. Thus, with \textit{``One-Time SVD''} performed offline for each model, \textit{``All-Time Decomposition''} becomes possible with minimal additional cost.

SVD-Cache delivers robust, near-lossless accelerations across models and tasks. On FLUX and HunyuanVideo, it achieves \textbf{5.55}\ $\times$ speedup with negligible quality drop. Moreover, it remains compatible with distillation, quantization and sparse-attention, with up to 29.01 $\times$ speedup on FLUX.1-schnell, 7.61$\times$ speedup on FLUX.1-DEV-int8 and 10.73$\times$ with sparse-attention under minimal degradation. Our contributions are summarized as follows:

\begin{itemize}
    \item \textbf{Subspace Heterogeneity.} We show that DiT features exhibit heterogeneous temporal dynamics across subspaces: the principal subspace evolves smoothly and predictably, while the residual is low-energy yet oscillatory and suitable for direct reuse. Through decomposition analyses across different prompts, we further reveal that the singular values and right singular matrices are input invariant.
    \item \textbf{SVD-Cache Framework.} Motivated by the observation that diffusion features evolve in a highly structured yet heterogeneous manner across different subspaces, we propose an SVD-based feature caching framework that partitions diffusion features into a rank-$k$ principal subspace where we apply EMA and a residual subspace directly reused.
    \item \textbf{Outstanding Results.} We validate SVD-Cache on multiple DiT models for both image and video generation, including FLUX and HunyuanVideo. Other acceleration models or methods such as FLUX.1-schnell, FLUX.1-Int8 and sparse attention were also evaluated. Our method consistently achieves state-of-the-art performance.
\end{itemize}
\section{Related Work}\label{sec:related-work}

Diffusion models~\citep{sohl2015deep,ho2020DDPM} have demonstrated remarkable capabilities in image and video generation. While early implementations commonly relied on U-Net architectures~\citep{ronneberger2015unet}, their scalability constraints were later mitigated by Diffusion Transformers (DiT)~\citep{peebles2023dit}, which enabled significant improvements in quality and resolution across multiple modalities~\citep{chen2023pixartalpha,chen2024pixartsigma,opensora,yang2025cogvideox}. Despite these advances, the core limitation of diffusion models remains their iterative sampling process, which introduces substantial computational overhead during inference. This has motivated two key research directions: minimizing the number of sampling steps, and reducing the per-step computational cost. Beyond raw efficiency, an ongoing challenge lies in preserving generative fidelity and stability, particularly under aggressive acceleration.

\subsection{Sampling Timestep Reduction}
A major approach to acceleration involves reducing the number of diffusion steps. DDIM~\citep{songDDIM} first introduced a deterministic sampling strategy that allows fast generation without compromising perceptual quality. Subsequently, higher-order solvers such as DPM-Solver and its variants~\citep{lu2022dpm,lu2022dpm++,zheng2023dpmsolvervF} have achieved better trade-offs between accuracy and efficiency by employing advanced numerical discretization schemes with controlled local truncation error. Other methods like Rectified Flow~\citep{refitiedflow} shorten transport trajectories in latent space, while progressive distillation~\citep{salimans2022progressive} compresses long diffusion chains into compact generative models. More recently, consistency models~\citep{song2023consistency} have enabled direct few-step synthesis by learning a noise-to-signal mapping in a consistent framework.

\subsection{Denoising Network Acceleration}
A wide range of model compression techniques aim to reduce the cost of each denoising step. These include structural pruning~\citep{structural_pruning_diffusion,zhu2024dipgo}, low-bit quantization~\citep{10377259,shang2023post,kim2025ditto}, knowledge distillation~\citep{li2024snapfusion}, and token-level reduction strategies~\citep{bolya2023tomesd,kim2024tofu,zhang2024tokenpruningcachingbetter,zhang2025sito,cheng2025catpruningclusterawaretoken}. Although these methods are effective in decreasing inference latency and memory usage, they often require additional retraining and may suffer performance degradation under distributional shifts.

Feature caching is an orthogonal acceleration paradigm that reduces redundant computation by reusing intermediate activations across timesteps. Early efforts in U-Net-style models~\citep{li2023FasterDiffusion,ma2024deepcache} have evolved into DiT-specific designs, including FasterCache~\citep{lv2025fastercachetrainingfreevideodiffusion}, FORA~\citep{selvaraju2024fora}, $\Delta$-DiT~\citep{chen2024delta-dit}, TeaCache~\citep{liu2024timestep}. These approaches vary in cache granularity and update frequency. Recent developments further extend this line of work through dynamic update schemes (e.g., ToCa, DuCa)~\citep{zou2024accelerating,zou2024DuCa}, unified cache-prune frameworks~\citep{sun2025unicpunifiedcachingpruning}, and region-aware sampling~\citep{liu2025regionadaptivesamplingdiffusiontransformers}. Notably, TaylorSeer~\citep{liu2025reusingforecastingacceleratingdiffusion} exemplifies the \textit{cache-then-forecast} paradigm, leveraging polynomial extrapolation over cached feature sequences to anticipate future activations with minimal overhead.

\section{Method}

\begin{figure*}[h]
    \centering
  \includegraphics[trim=160 230 308 140, clip,width=1\linewidth]{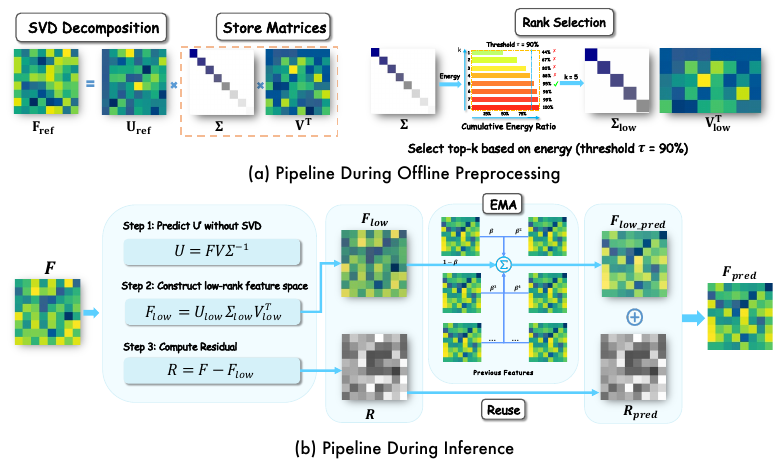}
    \caption{Overview of the proposed SVD-Cache framework. At each diffusion timestep, we decompose the feature map into principal and residual subspaces via SVD. The principal subspace, which captures the most significant variations and evolves smoothly, is predicted using exponential moving average (EMA). The residual subspace, containing high-frequency details that resist accurate prediction, is directly reused. This hybrid strategy allows us to efficiently cache features while maintaining high fidelity in the generated outputs.}
    \label{fig:method_overview}
\end{figure*}

\subsection{Preliminary}

\noindent \textbf{Diffusion Models.} Diffusion models generate data by simulating a forward–reverse stochastic process. The forward process gradually corrupts a data sample \(x_0\) with Gaussian noise:
\begin{equation}
    x_t = \sqrt{\alpha_t} x_0 + \sqrt{1 - \alpha_t} \, \epsilon_t, \quad \epsilon_t \sim \mathcal{N}(0, \mathbf{I}),
\end{equation}
where \(\alpha_t\) is a decreasing noise schedule. After \(T\) steps, the sample approaches pure noise. The reverse process learns to recover \(x_0\) by predicting the noise component \(\epsilon_\theta(x_t, t)\), commonly parameterized as:
\begin{equation}
    x_{t-1} = \frac{1}{\sqrt{\alpha_t}} \left( x_t - \frac{1 - \alpha_t}{\sqrt{1 - \bar{\alpha}_t}} \, \epsilon_\theta(x_t, t) \right) + \sigma_t \, \epsilon,
\end{equation}
where \(\bar{\alpha}_t = \prod_{s=1}^t \alpha_s\), and \(\sigma_t\) controls residual noise.

\noindent \textbf{Feature Caching.} Feature caching seeks to accelerate diffusion sampling by reusing intermediate activations across timesteps, thus avoiding redundant computation. At timestep \(t\), the network produces hidden feature maps \(\mathcal{F}_t = \{ \mathcal{F}_t^l \}_{l=1}^{L}\). A caching function \(\mathcal{C}(\mathcal{F}_A, k)\) estimates the feature \(\tilde{\mathcal{F}}_k\) at a future timestep \(k\) based on previously cached representations:
\begin{equation}
\tilde{\mathcal{F}}_k = \mathcal{C}(\mathcal{F}_t, k) := \mathcal{F}_t, \quad \forall k \in (t, t+n-1].
\end{equation}
This reuse strategy provides an approximate \((n{-}1)\times\) reduction in computation. Recent methods enhance this paradigm by forecasting future features rather than directly copying them.

\subsection{Low-Rank Approximation via SVD}
\label{sec:svd_lowrank}

Let $F\in \mathbb{R}^{N\times D}$ denote the feature matrix for a chosen DiT block. 
We compute a singular value decomposition (SVD):
\begin{equation}
\label{eq:svd}
F \;=\; U\,\Sigma\,V^{\top}\quad
U\in \mathbb{R}^{N\times r},\;\Sigma\in \mathbb{R}^{r\times r},\;V\in \mathbb{R}^{D\times r}
\end{equation}
where $r=\mathrm{rank}(F)$ and singular values in $\Sigma$ are sorted in descending order.
We retain the top-$k$ components ($k\!\ll\!r$) to form a rank-$k$ approximation:
\begin{equation}
\label{eq:trunc_svd}
F_k = U_{k}\,\Sigma_{k}\,V_{k}^{\top}.
\end{equation}

\noindent \textbf{Rank selection.}
We determine $k$ using an energy threshold $\tau$ based on the cumulative singular value energy:
\begin{equation}
\label{eq:energy_threshold}
\frac{\sum_{i=1}^k \sigma_i^2}{\sum_{i=1}^r \sigma_i^2} \;\ge\; \tau,
\end{equation}
where $\{\sigma_i\}$ are singular values of $F_t$.
This adaptive criterion preserves most of the representational energy while discarding redundant or noisy dimensions.

\subsection{One-Time SVD and Basis Reuse}
\label{sec:basis_reuse}

As observed in Fig.~\ref{fig:intro}(b), both the singular values $\sigma$ and the right singular matrix \(V\) exhibit minimal variation across different input prompts, which allows us to reuse these components for efficient low-rank subspace reconstruction.

\noindent \textbf{Reference Basis Caching.}
We first perform SVD on the feature space of a reference prompt and store its essential components for reuse. 
Specifically, we cache the right singular matrix \(V_{\mathcal{C}}\) and the singular value vector \(\bm{\sigma}_{\mathcal{C}}\):
\begin{equation}
\label{eq:cached_basis}
\textstyle \mathrm{cache}:\quad 
V_{\mathcal{C}} \in \mathbb{R}^{D\times r}, \quad 
\bm{\sigma}_{\mathcal{C}} = [\sigma_1, \ldots, \sigma_r]^{\top} \in \mathbb{R}^{r}.
\end{equation}

\noindent \textbf{Low-Rank Subspace Reconstruction.}
For any subsequent prompt with feature space \(F\), we reuse the cached basis to reconstruct its low-rank subspace without recomputing SVD.
The left singular matrix is approximated as:
\begin{equation}
\label{eq:uprime}
U = F V_{\mathcal{C}} \,\mathrm{diag}(\bm{\sigma}_{\mathcal{C}})^{-1}
\end{equation}
Then, the rank-\(k\) approximation of \(F\) is obtained by truncating the components:
\begin{equation}
\label{eq:fk_approx}
F_k = U_k\,\mathrm{diag}(\bm{\sigma}_{\mathcal{C},k})\,V_{\mathcal{C},k}^{\top}
\end{equation}
where \(U_k\), \(\bm{\sigma}_{\mathcal{C},k}\), and \(V_{\mathcal{C},k}\) denote the top-\(k\) components of \(U\), \(\bm{\sigma}_{\mathcal{C}}\), and \(V_{\mathcal{C}}\), respectively.

Finally, the residual component is computed as:
\begin{equation}
\label{eq:residual}
R = F - F_k
\end{equation}

\subsection{Exponential Moving Average for Prediction}
\label{sec:ema}

Exponential moving average (EMA) is a lightweight temporal forecasting method that assigns exponentially decaying weights to past observations. 
In our setting, we use EMA to predict the future low-rank features $\{F_{k,t}\}$ based on their stable temporal evolution across denoising steps.

\noindent \textbf{EMA Formulation.}
Let $F_{k,t}$ denote the rank-$k$ feature representation at caching timestep $t$. 
We maintain an EMA state $\widehat{F}_{k,t}$, which is recursively updated as:
\begin{equation}
\label{eq:ema_basic}
\widehat{F}_{k,t}
=
\beta\, \widehat{F}_{k,t-\Delta}
+
(1-\beta)\, F_{k,t},
\end{equation}
where $\beta \in (0,1)$ controls the trade-off between smoothness and responsiveness: larger $\beta$ yields smoother long-range predictions, while smaller $\beta$ reacts more quickly to short-term changes in $F_{k,t}$. The updated EMA state $\widehat{F}_{k,t}$ is then used to predict the next cached timestep, approximating $F_{k,t+\Delta}$. Practically, we set $\beta=0.9$ in all experiments.

\subsection{SVD-Cache Framework}
\label{sec:framework}

SVD-Cache establishes a shared low-rank basis by performing one-time SVD on a reference prompt. The resulting right singular vectors \(V_{\mathcal{C}}\) and singular values \(\boldsymbol{\sigma}_{\mathcal{C}}\) are cached following Eq.~\eqref{eq:cached_basis}, forming a universal subspace that can be reused across prompts. Given a new prompt with feature matrix \(F\), we reconstruct its low-rank representation in this cached subspace. The feature is projected onto the reference basis to obtain approximate left singular vectors using Eq.~\eqref{eq:uprime}, which are then truncated to produce the rank-\(k\) approximation \(F_k\) as in Eq.~\eqref{eq:fk_approx}. The orthogonal residual \(R\) is computed through Eq.~\eqref{eq:residual}; it captures the high-frequency, low-energy components that lie outside the principal subspace.

We then treat the subspaces respectively. The low-rank feature \(F_k\), which exhibits smooth and stable temporal evolution, is predicted using an EMA-based forecaster, yielding the next-step approximation \(\widehat{F}_{k,t+\Delta}\). The residual, in contrast, shows much weaker temporal coherence and is difficult to forecast reliably (as illustrated in Fig.~\ref{fig:intro}(a)). To prevent error amplification, we simply reuse it across timesteps: $\widehat{R}_{t+\Delta} = R_t.$The final feature at the future timestep is reconstructed by combining the EMA-predicted low-rank feature with the reused residual:
$\widehat{F}_{t+\Delta}=\widehat{F}_{k,t+\Delta}+\widehat{R}_{t+\Delta}.$

\begin{figure*}[t]
  \centering
  \includegraphics[trim=60 125 50 25, clip,width=0.95\linewidth]{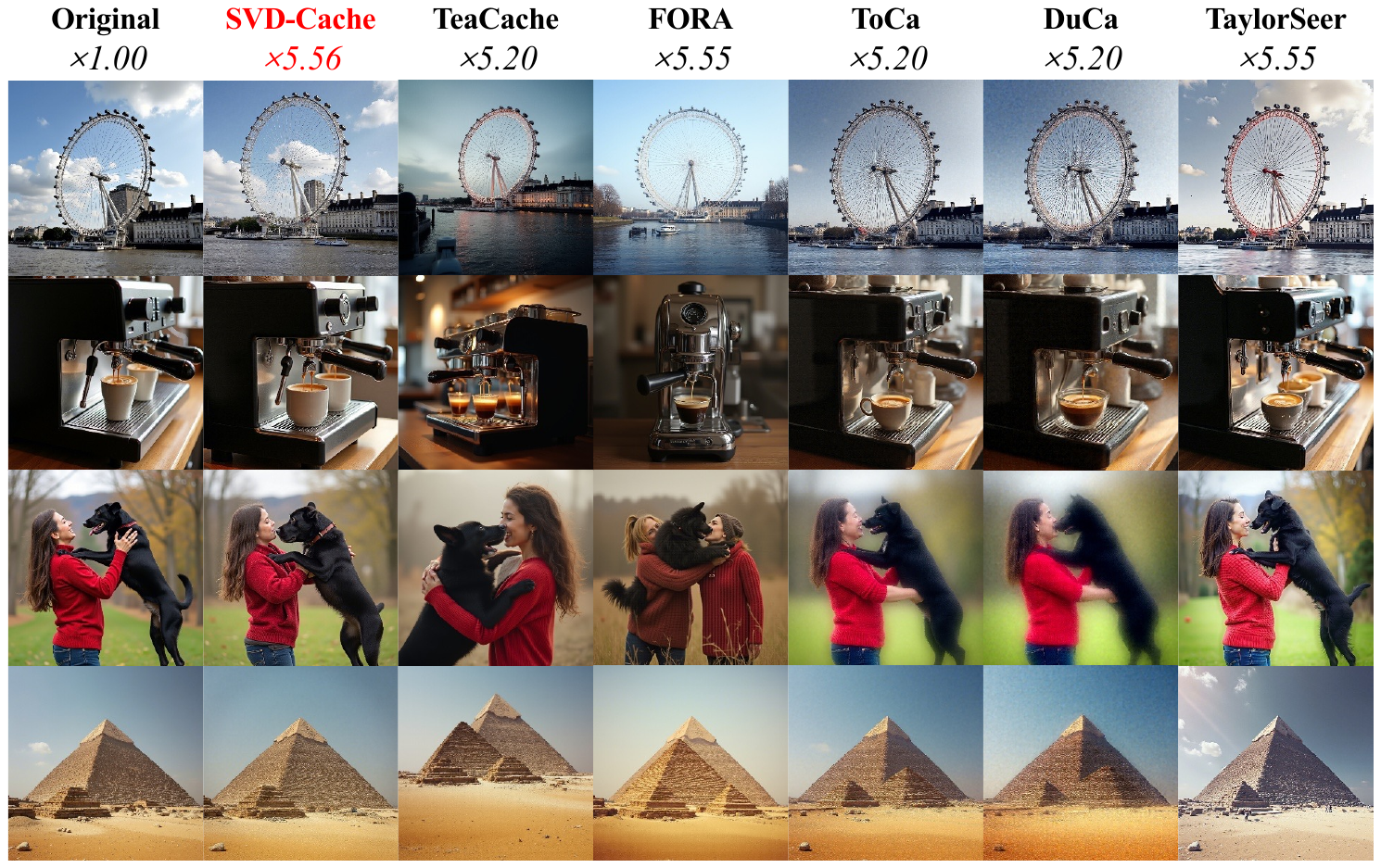}
  \caption{Visual Comparison of 5.5 $\times$ accelerated FLUX between different feature cache methods.}
  \label{fig:flux}
\end{figure*}

This hybrid design leverages the predictability of principal directions while safely reusing the high-frequency residual, enabling SVD-Cache to maintain accuracy and stability at significantly reduced computational cost.

\section{Experiments}

\subsection{Experiment Settings}

\begin{table*}[htbp]
\centering

\caption{\textbf{Quantitative comparison of text-to-image generation} on FLUX.1-dev.}

\setlength\tabcolsep{5pt} 
\small
\resizebox{\textwidth}{!}{
\begin{tabular}{l | c | c  c | c  c | c | c }
    \toprule
    \rowcolor{white}
  {\bf Method} & {\bf Efficient} & \multicolumn{4}{c|}{\bf Acceleration} & \multirow{2}{*} {\bf Image Reward $\uparrow$} & \multirow{2}{*}{\bf CLIP Score $\uparrow$} \\
    \cline{3-6}
    {\bf FLUX.1} & {\bf Attention } & {\bf Latency(s) $\downarrow$} & {\bf Speed $\uparrow$} & {\bf FLOPs(T) $\downarrow$}  & {\bf Speed $\uparrow$}\rule{0pt}{2ex} &  & \\
    \midrule
        $\textbf{[dev]: 50 steps}$ & \ding{52} & 25.82 & 1.00$\times$ & 3719.50 & 1.00$\times$ & 0.9898 \textcolor{gray!70}{\scriptsize (+0.000\%)} & 32.404 \textcolor{gray!70}{\scriptsize (+0.000\%)} \\ 
        \midrule

        {$60\%$\textbf{ steps}} & \ding{52} & 16.70 & 1.55$\times$ & 2231.70 & 1.67$\times$ & 0.9663 \textcolor{gray!70}{\scriptsize (-2.371\%)} & 32.312 \textcolor{gray!70}{\scriptsize (-0.283\%)} \\
        {$\Delta$-DiT} ($\mathcal{N}=2$) & \ding{52} & 17.80 & 1.45$\times$ & 2480.01 & 1.50$\times$ & 0.9444 \textcolor{gray!70}{\scriptsize (-4.594\%)} & 32.273 \textcolor{gray!70}{\scriptsize (-0.404\%)} \\
        {$\Delta$-DiT} ($\mathcal{N}=3$) & \ding{52} & 13.02 & 1.98$\times$ & 1686.76 & 2.21$\times$ & 0.8721 \textcolor{gray!70}{\scriptsize (-11.90\%)} & 32.102 \textcolor{gray!70}{\scriptsize (-0.933\%)} \\
        $\textbf{TaylorSeer}$ $(\mathcal{N}=3, O=2)$ & \ding{52} & 9.89 & 2.61$\times$ & 1320.07 & 2.82$\times$ & 0.9989 \textcolor{gray!70}{\scriptsize (+0.919\%)} & 32.413 \textcolor{gray!70}{\scriptsize (+0.027\%)} \\
        $\textbf{FoCa}$ $(\mathcal{N}=3)$ & \ding{52} & 9.28 & 2.78$\times$ & 1327.21 & 2.80$\times$ & 0.9890 \textcolor{gray!70}{\scriptsize (-0.081\%)} & 32.577 \textcolor{gray!70}{\scriptsize (+0.533\%)} \\
        
        \rowcolor{gray!20}
        $\textbf{SVD-Cache}$ $(\mathcal{N}=4)$ & \ding{52} & \bf 8.09 & \textbf{3.19$\times$} & \bf 967.91 & \textbf{3.84$\times$} &\bf 1.0147 \textcolor[HTML]{0f98b0}{\scriptsize (+2.519\%)} & \bf 32.840 \textcolor[HTML]{0f98b0}{\scriptsize (+1.348\%)} \\
        \midrule
        
        {$34\%$\textbf{ steps}} & \ding{52} & 9.07 & 2.85$\times$ & 1264.63 & 3.13$\times$ & 0.9453 \textcolor{gray!70}{\scriptsize (-4.498\%)} & 32.114 \textcolor{gray!70}{\scriptsize (-0.893\%)} \\
        $\textbf{Chipmunk}$ & \ding{52} & 12.72 & 2.02$\times$ & 1505.87 & 2.47$\times$ & 0.9936 \textcolor{gray!70}{\scriptsize (+0.384\%)} & 32.548 \textcolor{gray!70}{\scriptsize (+0.444\%)} \\
        $\textbf{FORA}$ $(\mathcal{N}=3)$ & \ding{52} & 10.16 & 2.54$\times$ & 1320.07 & 2.82$\times$ & 0.9776 \textcolor{gray!70}{\scriptsize (-1.232\%)} & 32.266 \textcolor{gray!70}{\scriptsize (-0.425\%)} \\
        $\textbf{\texttt{ToCa}}$ $(\mathcal{N}=6)$ & \ding{56} & 13.16 & 1.96$\times$ & 924.30 & 4.02$\times$ & 0.9802 \textcolor{gray!70}{\scriptsize (-0.968\%)} & 32.083 \textcolor{gray!70}{\scriptsize (-0.990\%)} \\
        $\textbf{\texttt{DuCa}}$ $(\mathcal{N}=5)$ & \ding{52} & 8.18 & 3.15$\times$ & 978.76 & 3.80$\times$ & 0.9955 \textcolor{gray!70}{\scriptsize (+0.576\%)} & 32.241 \textcolor{gray!70}{\scriptsize (-0.503\%)} \\
        $\textbf{TaylorSeer}$ $(\mathcal{N}=4, O=2)$ & \ding{52} & 9.24 & 2.80$\times$ & 967.91 & 3.84$\times$ & 0.9857 \textcolor{gray!70}{\scriptsize (-0.414\%)} & 32.413 \textcolor{gray!70}{\scriptsize (+0.027\%)} \\
        $\textbf{FoCa}$ $(\mathcal{N}=4)$ & \ding{52} & 9.35 & 2.76$\times$ & 1050.70 & 3.54$\times$ & 0.9757 \textcolor{gray!70}{\scriptsize (-1.424\%)} & 32.538 \textcolor{gray!70}{\scriptsize (+0.414\%)} \\
        $\textbf{Clusca}$ $(\mathcal{N}=4, O=2, K=16)$ & \ding{52} & 9.25 & 2.79$\times$ & 1045.58 & 3.56$\times$ & 0.9850 \textcolor{gray!70}{\scriptsize (-0.485\%)} & 32.441 \textcolor{gray!70}{\scriptsize (+0.114\%)} \\
        \rowcolor{gray!20}
        $\textbf{SVD-Cache}$ $(\mathcal{N}=5)$ & \ding{52} & \bf 7.62 & \textbf{3.38$\times$} & \bf 893.54 & \textbf{4.16$\times$} &\bf 1.0123 \textcolor[HTML]{0f98b0}{\scriptsize (+2.267\%)} &\bf 32.983 \textcolor[HTML]{0f98b0}{\scriptsize (+1.788\%)} \\
        \midrule
        
        $\textbf{FORA}$ $(\mathcal{N}=4)$ & \ding{52} & 8.12 & 3.14$\times$ & 967.91 & 3.84$\times$ & 0.9730 \textcolor{gray!70}{\scriptsize (-1.695\%)} & 32.142 \textcolor{gray!70}{\scriptsize (-0.809\%)} \\
        $\textbf{\texttt{ToCa}}$ $(\mathcal{N}=8)$ & \ding{56} & 11.36 & 2.27$\times$ & 784.54 & 4.74$\times$ & 0.9451 \textcolor{gray!70}{\scriptsize (-4.514\%)} & 31.993 \textcolor{gray!70}{\scriptsize (-1.271\%)} \\
        $\textbf{\texttt{DuCa}}$ $(\mathcal{N}=7)$ & \ding{52} & \bf6.74 & \textbf{3.83$\times$} & 760.14 & 4.89$\times$ & 0.9757 \textcolor{gray!70}{\scriptsize (-1.424\%)} & 32.066 \textcolor{gray!70}{\scriptsize (-1.046\%)} \\
        \textbf{TeaCache} $({l}=0.8)$ & \ding{52} & 7.21 & 3.58$\times$ & 892.35 & 4.17$\times$ & 0.8683 \textcolor{gray!70}{\scriptsize (-12.28\%)} & 31.704 \textcolor{gray!70}{\scriptsize (-2.159\%)} \\
        $\textbf{TaylorSeer}$ $(\mathcal{N}=5, O=2)$ & \ding{52} & 7.46 & 3.46$\times$ & 893.54 & 4.16$\times$ & 0.9768 \textcolor{gray!70}{\scriptsize (-1.314\%)} & 32.467 \textcolor{gray!70}{\scriptsize (+0.194\%)} \\
        $\textbf{FoCa}$ $(\mathcal{N}=6)$ & \ding{52} & 7.54 & 3.42$\times$ & 745.39 & 4.99$\times$ & 0.9713 \textcolor{gray!70}{\scriptsize (-1.870\%)} & {32.922} \textcolor{gray!70}{\scriptsize (+1.600\%)} \\
        $\textbf{Speca}$ $(\mathcal{N}_{\text{max}}=8, \mathcal{N}_{\text{min}}=2)$ & \ding{52} & 7.42 & 3.48$\times$ & 791.38 & 4.70$\times$ & 0.9985 \textcolor{gray!70}{\scriptsize (+0.878\%)} & 32.277 \textcolor{gray!70}{\scriptsize (-0.391\%)} \\
        $\textbf{Clusca}$ $(\mathcal{N}=5, O=1, K=16)$ & \ding{52} & 7.05 & 3.66$\times$ & 897.03 & 4.14$\times$ & 0.9718 \textcolor{gray!70}{\scriptsize (-1.818\%)} & 32.319 \textcolor{gray!70}{\scriptsize (-0.262\%)} \\
        \rowcolor{gray!20}
        $\textbf{SVD-Cache}$ $(\mathcal{N}=6)$ & \ding{52} &  6.81 & {3.79$\times$} & \bf 744.81 & \textbf{5.00$\times$} &\bf 1.0057 \textcolor[HTML]{0f98b0}{\scriptsize (+1.605\%)} & \bf33.027 \bf \textcolor[HTML]{0f98b0}{\scriptsize (+1.921\%)} \\
        \midrule

        $\textbf{FORA}$ $(\mathcal{N}=6)$ & \ding{52} & 8.17 & 3.16$\times$ & 744.80 & 4.99$\times$ & 0.7760 \textcolor{gray!70}{\scriptsize (-21.62\%)} & 31.742 \textcolor{gray!70}{\scriptsize (-2.043\%)} \\
        $\textbf{\texttt{ToCa}}$ $(\mathcal{N}=10)$ & \ding{56} & 7.93 & 3.25$\times$ & 714.66 & 5.20$\times$ & 0.7155 \textcolor{gray!70}{\scriptsize (-27.70\%)} & 31.808 \textcolor{gray!70}{\scriptsize (-1.839\%)} \\
        $\textbf{\texttt{DuCa}}$ $(\mathcal{N}=9)$ & \ding{52} & 7.27 & 3.55$\times$ & 690.25 & 5.39$\times$ & 0.8382 \textcolor{gray!70}{\scriptsize (-15.33\%)} & 31.759 \textcolor{gray!70}{\scriptsize (-1.993\%)} \\
        \textbf{TeaCache} $({l}=1)$ & \ding{52} & 8.19 & 3.19$\times$ & 743.63 & 5.01$\times$ & 0.8379 \textcolor{gray!70}{\scriptsize (-15.36\%)} & 31.877 \textcolor{gray!70}{\scriptsize (-1.627\%)} \\
        $\textbf{TaylorSeer}$ $(\mathcal{N}=7, O=2)$ & \ding{52} & 6.77 & 3.81$\times$ & 671.39 & 5.54$\times$ & 0.9698 \textcolor{gray!70}{\scriptsize (-2.020\%)} & 32.128 \textcolor{gray!70}{\scriptsize (-0.851\%)} \\
        $\textbf{Clusca}$ $(\mathcal{N}=6, O=1, K=16)$ & \ding{52} & 7.13 & 3.62$\times$ & 748.48 & 4.97$\times$ & 0.9704 \textcolor{gray!70}{\scriptsize (-1.956\%)} & 32.217 \textcolor{gray!70}{\scriptsize (-0.577\%)} \\
        \rowcolor{gray!20}
        $\textbf{SVD-Cache}$ $(\mathcal{N}=7)$ & \ding{52} &  6.43 & {4.01$\times$} &  670.44 & {5.55$\times$} &\bf 0.9938 \textcolor[HTML]{0f98b0}{\scriptsize (+0.404\%)} &\bf 33.144 \textcolor[HTML]{0f98b0}{\scriptsize (+2.282\%)} \\
        \rowcolor{gray!20}
        $\textbf{SVD-Cache}$ $(\mathcal{N}=8)$ & \ding{52} & \bf 4.99 & \textbf{5.17$\times$} & \bf 596.07 & \textbf{6.24$\times$} & 0.9769 \bf \textcolor[HTML]{0f98b0}{\scriptsize (-1.306\%)} & 32.848 \bf \textcolor[HTML]{0f98b0}{\scriptsize (+1.369\%)} \\
\bottomrule
\end{tabular}}
\label{table:FLUX}
\footnotesize
\end{table*}

\begin{table*}[htb]
\centering

\caption{\textbf{Quantitative comparison of text-to-video generation} on HunyuanVideo.}

\setlength\tabcolsep{8.0pt} 
\small
\resizebox{\textwidth}{!}{
\begin{tabular}{l | c | c  c | c  c | c }
    \toprule
    {\bf Method} & {\bf Efficient} &\multicolumn{4}{c|}{\bf Acceleration} &{\bf VBench $\uparrow$}  \\
    \cline{3-6}
    {\bf HunyuanVideo} & {\bf Attention } & {\bf Latency(s) $\downarrow$} & {\bf Speed $\uparrow$} & {\bf FLOPs(T) $\downarrow$}  & {\bf Speed $\uparrow$} & \bf Score(\%)\rule{0pt}{2ex}\\ 
    \midrule

  $\textbf{Original: 50 steps}$ 
                           & \ding{52}  & 145.00 & 1.00$\times$& {29773.0}   & {1.00$\times$} & 80.66 \textcolor{gray!70}{\scriptsize (+0.0\%)}      \\ 
    \midrule
  
  {$22\%$\textbf{ steps}}  & \ding{52}  & 31.87 & 4.55$\times$ & {6550.1}   & {4.55$\times$} & 78.74 \textcolor{gray!70}{\scriptsize (-2.4\%)}          \\

\textbf{AdaCache}
                           & \ding{52} & 55.34 & 2.62$\times$ & 11192.8 & 2.66$\times$ & 80.12 \textcolor{gray!70}{\scriptsize (-0.7\%)}  \\

\textbf{TeaCache}$({l}=0.4)$ 
                           & \ding{52} & 30.49 & 4.76$\times$ & 6550.1 & 4.55$\times$ & 79.36 \textcolor{gray!70}{\scriptsize (-1.6\%)}  \\

$\textbf{FORA}$ $(\mathcal{N}=6)$
& \ding{52}  & 34.39 & 4.22$\times$  & {5960.4}   & {5.00$\times$} & 78.83 \textcolor{gray!70}{\scriptsize (-2.3\%)}     \\

$\textbf{\texttt{ToCa}}$ $(\mathcal{N}=5,R=90\%)$
                           & \ding{56}  & 38.52 & 3.76$\times$ & {7006.2}   & {4.25$\times$} & 78.86 \textcolor{gray!70}{\scriptsize (-2.2\%)}    \\
  
$\textbf{\texttt{DuCa}} $ $(\mathcal{N}=5,R=90\%)$ 
                           & \ding{52}  & 31.69 & 4.58$\times$ & {6483.2}   & {4.48$\times$} & 78.72 \textcolor{gray!70}{\scriptsize (-2.4\%)}    \\
                           
$\textbf{TaylorSeer} $ $(\mathcal{N}=5,O=1)$ 
                           & \ding{52}  & 34.84 & 4.16$\times$ & {5960.4}  & {5.00$\times$} & 79.93 \textcolor{gray!70}{\scriptsize (-0.9\%)} \\

$\textbf{Speca} $ $(\mathcal{N}_{\text{max}}=8, \mathcal{N}_{\text{min}}=2)$ 
                           & \ding{52}  & 34.58 & 4.19$\times$ & {5692.7}  & {5.23$\times$} & 79.98 \textcolor{gray!70}{\scriptsize (-0.8\%)} \\

$\textbf{Clusca} $ $(\mathcal{N}=5,O=1,K=16)$ 
                           & \ding{52}  & 35.37 & 4.10$\times$ & {5373.0}  & {5.54$\times$} & 79.96 \textcolor{gray!70}{\scriptsize (-0.9\%)} \\

$\textbf{FoCa} $ $(\mathcal{N}=5)$ 
                           & \ding{52}  & 34.52 & 4.20$\times$ & {5966.5}  & {4.99$\times$} & 79.96 \textcolor{gray!70}{\scriptsize (-0.8\%)} \\

\rowcolor{gray!20}
$\textbf{SVD-Cache} $ $(\mathcal{N}=5)$ 
                           & \ding{52}  & 34.69 & {4.18$\times$} & 5960.4  & {5.00$\times$} & \bf {80.60} \textcolor[HTML]{0f98b0}{\scriptsize (-0.1\%)}  \\

\rowcolor{gray!20}
$\textbf{SVD-Cache} $ $(\mathcal{N}=6)$ 
                           & \ding{52}  & \bf 29.77 & \textbf{4.87$\times$} & \bf 5359.1  & \bf {5.56$\times$} &  {80.46} \bf \textcolor[HTML]{0f98b0}{\scriptsize (-0.2\%)}  \\

    \bottomrule
\end{tabular}}
\label{table:HunyuanVideo-Metrics}
\end{table*}

\begin{table*}[htbp]
\centering
\caption{\textbf{Quantitative comparison of other accelerated models} on FLUX.}

\resizebox{\textwidth}{!}{
\begin{tabular}{l | c | c  c | c  c | c  c | c c c}
    \toprule
    \multirow{2}{*}{\textbf{Method}} 
    & \multirow{2}{*}{\textbf{NFE}} 
    & \multicolumn{4}{c|}{\textbf{Acceleration}} 
    & \multicolumn{2}{c|}{\textbf{Quality Metrics}} 
    & \multicolumn{3}{c}{\textbf{Perceptual Metrics}} \\
    \cline{3-11}
    & 
    & \textbf{Latency(s) \(\downarrow\)} 
    & \textbf{Speed \(\uparrow\)} 
    & \textbf{FLOPs(T) \(\downarrow\)}  
    & \textbf{Speed \(\uparrow\)} 
    & \textbf{ImageReward\(\uparrow\)} 
    & \textbf{CLIP\(\uparrow\)} 
    & \textbf{PSNR\(\uparrow\)} 
    & \textbf{SSIM\(\uparrow\)} 
    & \textbf{LPIPS\(\downarrow\)}\rule{0pt}{2ex} \\
    \midrule

\textbf{FLUX.1[dev]} 
& 50 & 25.82 & 1.00$\times$ & 3719.50 & 1.00$\times$ & 0.9898 & 32.40 & - & - & - \\

\midrule

\textbf{FLUX.1[dev]-int8}
& 50 & 13.17 & 1.96$\times$ & 1823.30 & 2.04$\times$ & 1.0160 & 32.60 & $\infty$ & 1.000 & 0.000 \\

\rowcolor{gray!20}
\textbf{SVD-Cache ($\mathcal{N}=5$)} 
& 50 & 3.39 & 7.61$\times$ & 438.29 & 8.49$\times$ & 0.9904 & \textbf{32.88} & \textbf{29.74} & \textbf{0.708} & \textbf{0.3214} \\

\midrule

\textbf{FLUX[SpargeAttention]}
& 50 & 7.36 & 3.51$\times$ & 979.50 & 3.80$\times$ &  0.8358 & 32.49 & $\infty$ & 1.000 & 0.000 \\

\rowcolor{gray!20}
\textbf{SVD-Cache ($\mathcal{N}=5$)} 
& 50 & 2.40 & 10.73$\times$ & 318.45 & 11.68$\times$ & 0.8177 & \textbf{32.48} & \textbf{30.02} & \textbf{0.808} & \textbf{0.254} \\

\midrule

\textbf{FLUX[schnell]}
& 8 & 4.21 & 6.14$\times$ & 595.12 & 6.25$\times$ & 0.9090 & 33.93 & $\infty$ & 1.000 & 0.000 \\

\rowcolor{gray!20}
\textbf{SVD-Cache ($\mathcal{N}=2$)} 
& 8 &  2.85&  9.06$\times$ &  372.15 & 9.99$\times$ & {0.9285} & {33.92} & \textbf{34.81} & \textbf{0.936} & \textbf{0.047} \\

\rowcolor{gray!20}
\textbf{SVD-Cache ($\mathcal{N}=5$)} 
& 8 & \bf 1.69 & \bf 15.22$\times$ & \bf 223.20& \bf 16.66$\times$ & \bf 0.9713 &\textbf{34.68} & {30.01} & {0.745} & {0.272} \\

\midrule

\textbf{FLUX[schnell]}
& 4 & 2.34 & 11.03$\times$ & 297.60 & 12.50$\times$ & 0.9129 & 34.15 & $\infty$ & 1.000 & 0.000 \\

\textbf{TaylorSeer} ($\mathcal{N}=2$)
& 4 & 1.58 & 16.34$\times$ & 209.70 & 17.74$\times$ & 0.9012 & 33.76 & 29.13 & 0.746 & 0.249 \\

\textbf{TeaCache} $({l}=0.6)$
& 4 & 1.26 & 20.49$\times$ & 163.78 & 22.71$\times$ & 0.8921 & 33.87 & 28.01 & 0.379 & 0.734 \\

\rowcolor{gray!20}
\textbf{SVD-Cache ($\mathcal{N}=2$)} 
& 4 & {1.16} & {22.25$\times$} &{152.32} & {24.42$\times$} & {0.9247} & {34.20} & \textbf{34.83} & \textbf{0.947} & \textbf{0.037} \\

\rowcolor{gray!20}
\textbf{SVD-Cache ($\mathcal{N}=3$)} 
& 4 & \textbf{0.89} & \textbf{29.01$\times$} & \textbf{74.58} & \textbf{49.87$\times$} & \bf {0.9463} & \textbf{34.65} & {29.70} & {0.785} & {0.176} \\

\bottomrule
\end{tabular}}

{\scriptsize
\begin{itemize}[leftmargin=10pt,topsep=0pt]
\item The PSNR, SSIM, and LPIPS of SVD-Cache are computed with respect to the outputs of the corresponding accelerated baselines instead of the original model. SpargeAttention\citep{zhang2024sageattention2} \citep{zhang2025sageattention} \citep{zhang2025spargeattn}  is a a universal training-free sparse attention accelerating language, image, and video models.
\end{itemize}
}

\label{table:dis}
\end{table*}

\noindent\textbf{Model Configurations.}
We evaluate two representative diffusion models—FLUX.1-dev~\citep{flux2024} for text-to-image and HunyuanVideo~\citep{sun_hunyuan-large_2024} for text-to-video—and further test compatibility with FLUX.1-dev-torchao-int8 (quantization), FLUX.1-schnell (step distillation), and sparse-attention variants. This setup verifies that our method integrates seamlessly with quantization, distillation, and sparse-attention–based acceleration techniques.

\noindent\textbf{Evaluation and Metrics.}
For text-to-image tasks, we follow DrawBench~\citep{saharia2022drawbench}, reporting ImageReward~\citep{xu2023imagereward}, CLIP Score~\citep{hessel2021clipscore}, and PSNR/SSIM/LPIPS. For text-to-video generation, we evaluate HunyuanVideo using VBench~\citep{VBench}, which measures motion, appearance, and semantic consistency.

\begin{figure}[htp]
  \centering
  \includegraphics[trim=73 165 80 84, clip,width=1\linewidth]{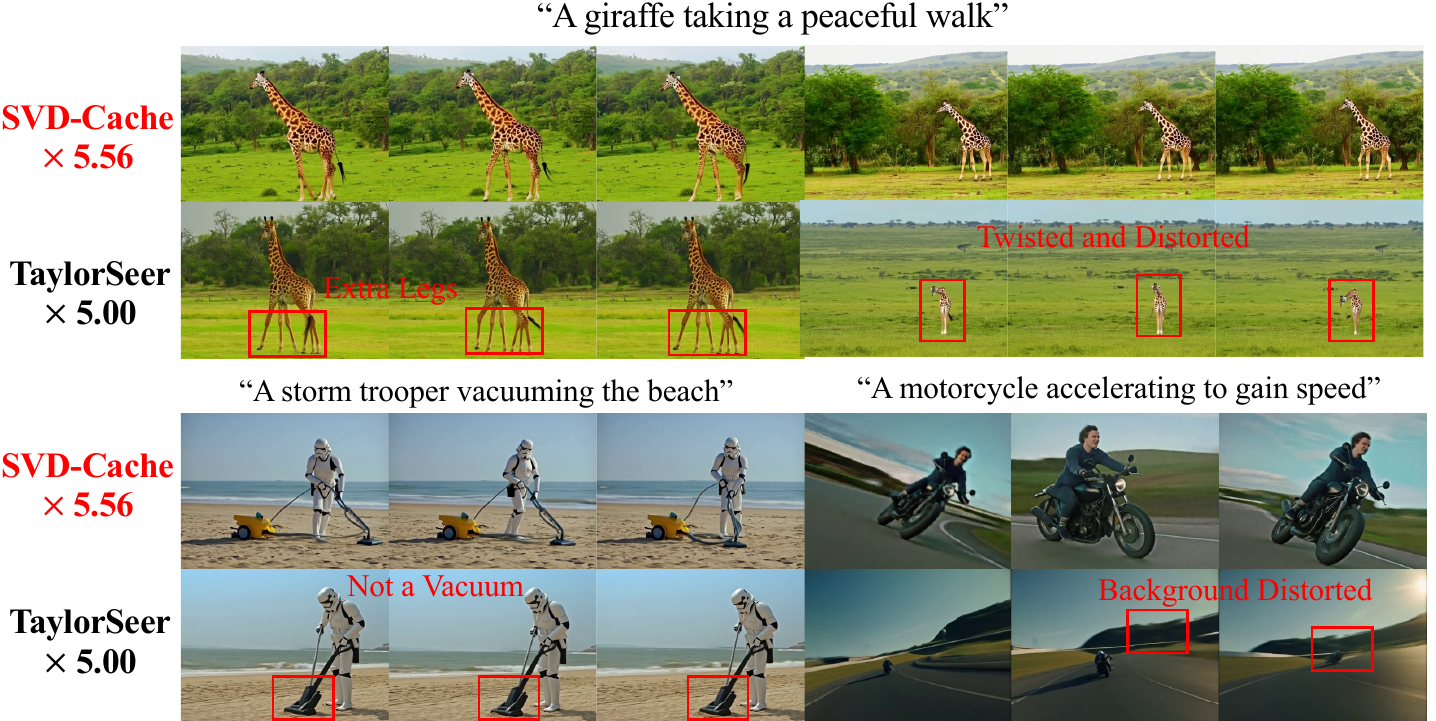}
  \caption{Comparison between SVD-Cache and TaylorSeer on HunyuanVideo. SVD-Cache sustains stronger temporal consistency and visual quality under more aggressive acceleration, while TaylorSeer develops visible artifacts and flickering.}
\end{figure}

\subsection{Results on Text-to-Image Generation}

As shown in Table~\ref{table:FLUX}, SVD-Cache consistently outperforms all baselines across acceleration levels on FLUX.1-dev. At interval \(\mathcal{N}=5\), it attains the best ImageReward (\textbf{1.0123}) and CLIP (\textbf{32.983}) with a 4.16$\times$ FLOPs reduction, surpassing reuse-based methods and prediction-based approaches such as TaylorSeer (0.9768 at 4.16$\times$) and FoCa (0.9713 at 4.99$\times$). At \(\mathcal{N}=6\), it remains competitive, achieving \textbf{1.0057} ImageReward and \textbf{33.027} CLIP at 5.00$\times$, outperforming TaylorSeer and Speca under similar cost. Even under aggressive compression, SVD-Cache remains robust: at \(\mathcal{N}=7\), it delivers \textbf{0.9938} ImageReward with 5.55$\times$ FLOPs (vs. TaylorSeer 0.9698, FoCa 0.9713); and at \(\mathcal{N}=8\), it still maintains strong quality (ImageReward \textbf{0.9769}, CLIP \textbf{32.848}) with a 6.24$\times$ speedup. These results highlight SVD-Cache's superior efficiency–quality trade-off across a broad range of compression regimes.

\subsection{Results on Text-to-Video Generation.}  

We evaluate SVD-Cache on HunyuanVideo and compare it with recent acceleration baselines. As shown in Table~\ref{table:HunyuanVideo-Metrics}, SVD-Cache achieves state-of-the-art performance across all acceleration settings. At \(\mathcal{N}=5\), it reaches a VBench score of \textbf{80.60}, nearly matching the original model (80.66) while reducing FLOPs by \textbf{5.00$\times$}, outperforming TaylorSeer (79.93), Speca (79.98), and FoCa (79.96) at similar cost. At \(\mathcal{N}=6\), it achieves the fastest latency (\textbf{29.77s}) and a \textbf{5.56$\times$} FLOPs reduction, while maintaining a strong VBench score of \textbf{80.46}. This surpasses robust baselines such as DuCa (78.72) and ToCa (78.86), demonstrating superior stability under aggressive skipping. Overall, these results confirm the robustness of SVD-Cache for video generation.

\subsection{Results on Acceleration Methods.}  
\begin{figure}[htp]
  \centering
  \includegraphics[trim=160 190 160 60, clip,width=1\linewidth]{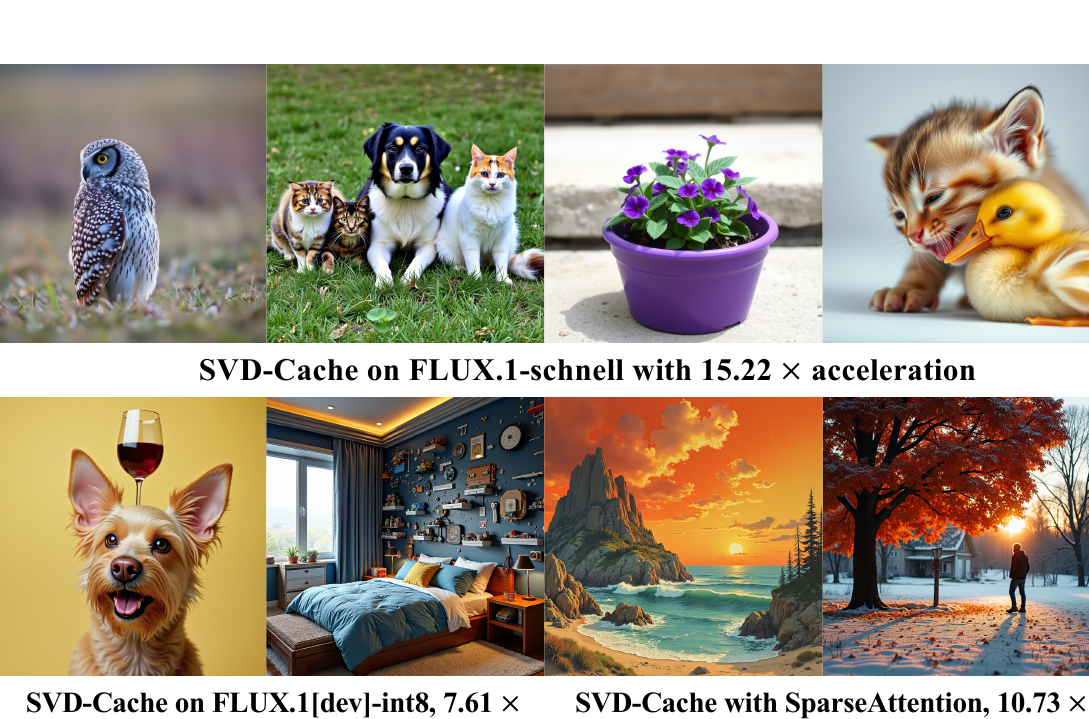}
  \caption{Images of SVD-Cache applied on different accelerated diffusion models.}
  \label{fig:acceleration_methods}
\end{figure}

To assess generality, we apply SVD-Cache to several accelerated diffusion models and methods: the INT8-quantized FLUX.1-dev-int8, the step-distilled FLUX.1-schnell and sparse attention. As summarized in Table~\ref{table:dis}, SVD-Cache consistently provides significant speedup while maintaining or improving quality. Visual examples are shown in Fig.~\ref{fig:acceleration_methods}.

On FLUX.1-dev-int8, SVD-Cache ($\mathcal{N}=5$) reaches an ImageReward of \textbf{0.9904} and CLIP \textbf{32.88}, and surpasses the original quantized baseline (1.0160) on PSNR (29.74), SSIM (0.708), and LPIPS (0.3214).On the distilled FLUX.1-schnell ($\mathrm{NFE}=4,8$), it offers flexible trade-offs—for example, at $\mathcal{N}=3$, it delivers a 49.87$\times$ FLOPs reduction with an ImageReward of \textbf{0.9463}, outperforming TaylorSeer and TeaCache under the same NFE.

SVD-Cache is orthogonal to quantization, distillation, and sparse attention, operating purely at inference through feature reuse and solver scheduling. It integrates seamlessly with existing compression methods, enabling additive speedups without retraining or reparameterization.
\begin{figure*}[t]
  \centering
  \includegraphics[trim=132 190 175 200, clip,width=1\linewidth]{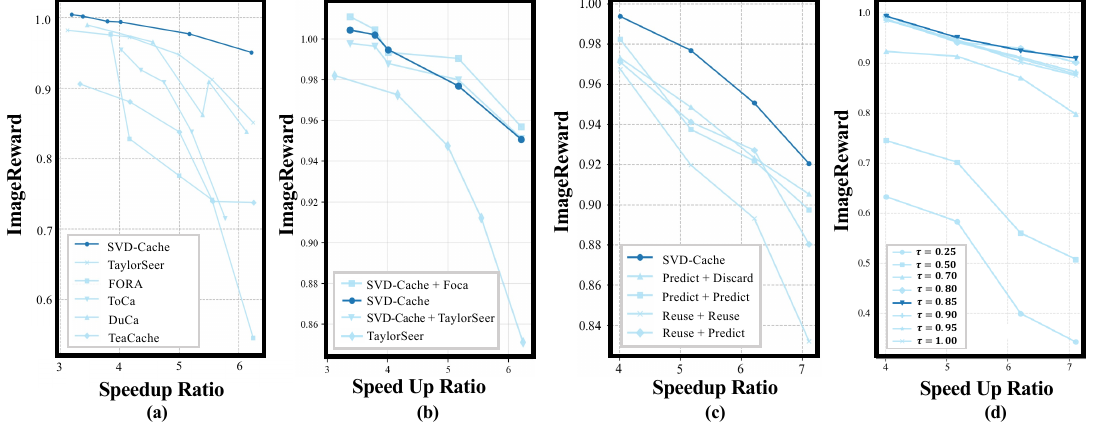}
  \caption{\textbf{Overall and Ablation Results of SVD-Cache.}~(a)~SVD-Cache consistently outperforms token-wise(ToCa) and other full-feature-space predictor baselines.~(b)~ When the low-rank subspace is predicted with other ODE methods, quality is significantly improved over the original method. ~(c)~Ablation study on Flux by applying different strategies to the low-rank and residual components separately.~(d)~Ablation study on the energy threshold $\tau$ for subspace decomposition.}
  \label{fig:OverallAblation}
\end{figure*}

\subsection{Ablation Study.}

Ablation study is conducted on FLUX. We first investigate the impact of applying different methods on the low-rank subspace and the residual component separately. As shown in Fig~\ref{fig:OverallAblation}(c), applying EMA prediction on the low-rank subspace while reusing the residual yields the best performance, validating our design choice. Using direct reuse for both components leads to significant quality degradation, while applying EMA to both results in error accumulation.

We also conduct an ablation study to investigate how different choices of the energy threshold $\tau$ (Eq(\ref{eq:energy_threshold})) affect the decomposition. As shown in Fig.~\ref{fig:OverallAblation}(d), we find that the optimal energy threshold is around 0.85. Setting $\tau$ higher leads to an overly large low-rank subspace that absorbs high-frequency noise, reducing prediction stability and degrading EMA accuracy.
Conversely, choosing a smaller $\tau$ makes the low-rank subspace too restrictive, discarding essential structural components and causing noticeable errors.

\section{Discussion}

\subsection{Why Feature Space Decomposition?}

Prior caching methods implicitly assume that all feature dimensions evolve smoothly over time, which is rarely true in high-dimensional hidden states. Only a part of feature space shows stable, predictable dynamics based on our observation; the remaining contains high-frequency fluctuations that are difficult to forecast and easily amplify prediction errors. Operating directly in the full space therefore mixes stable and unstable directions, leading to rapid error growth.

By decomposing features, SVD-Cache predicts only the smooth, low-rank component where forecasting is reliable, while handling the volatile residual conservatively through reuse or mild smoothing. This selective treatment makes caching both more stable and accurate. As shown in Fig.~\ref{fig:OverallAblation}(a), this decomposition consistently outperforms variants that attempt to predict the full feature space.

\subsection{Compatibility with Acceleration Methods.} 

SVD-Cache follows a central principle of diffusion acceleration: \textbf{preserve essential structure} while down-weighting components with limited influence. By decomposing features into a dominant subspace and an orthogonal residual, it forecasts only the stable, high-utility directions and handles the volatile tail conservatively—naturally aligning with major acceleration families.

Distilled models benefit because SVD-Cache avoids extrapolating along unstable, high-curvature directions where full-space predictors often fail. Quantized models gain from prediction in a compressed coordinate system that attenuates quantization noise, with the residual reused rather than aggressively forecasted. Sparse-attention methods benefit similarly, as SVD-Cache emphasizes principal structure in feature space. Together, subspace-restricted forecasting, residual stabilization, and one-time basis reuse provide robustness and low overhead, integrating cleanly with step-skipping or lightweight sampling schedules.
\subsection{Compatibility with Other Caching Methods}

SVD-Cache is orthogonal to existing caching methods such as TaylorSeer~\citep{liu2025reusingforecastingacceleratingdiffusion} and FoCa~\citep{zheng2025forecastcalibratefeaturecaching}, which focus on designing better predictors over the entire feature space. In contrast, SVD-Cache introduces a subspace-aware strategy that decomposes the feature space and applies different caching strategies to each subspace. Therefore, SVD-Cache can be integrated with these methods by replacing their full-space predictors with our subspace-aware design, yielding further gains in caching efficiency and quality preservation, as shown in Fig.~\ref{fig:OverallAblation}(b). When paired with TaylorSeer, SVD-Cache delivers substantial improvements over the TaylorSeer, and its combination with more recent approaches such as FoCa demonstrates even more promising results. 
\section{Conclusion}

We presented SVD-Cache, a training-free feature caching framework for diffusion transformers that leverages low-rank feature decomposition and stable subspace prediction to reduce computation while preserving generation quality. Through a one-time SVD and dimension-wise residual modeling, SVD-Cache provides robust, scalable caching across prompts and timesteps. Extensive experiments on images and videos demonstrate consistent state-of-the-art performance across a wide range of acceleration ratios, outperforming existing reuse and forecast-based methods. Moreover, SVD-Cache is orthogonal to quantization, distillation, and sparse attention, enabling additive speedups without retraining and providing a practical path toward scalable, plug-and-play acceleration for modern diffusion models.
\clearpage
{
    \small
    \bibliographystyle{ieeenat_fullname}
    \bibliography{main}

@String(ICCV= {Int. Conf. Comput. Vis.})

@String(ICLR = {Int. Conf. Learn. Represent.})

@String(AAAI = {AAAI})

@String(ICCV  = {ICCV})

@String(ICLR  = {ICLR})

@article{structural_pruning_diffusion,
  title        = {Structural Pruning for Diffusion Models},
  author       = {Fang, Gongfan and Ma, Xinyin and Wang, Xinchao},
  year         = 2023,
  journal      = {arXiv preprint arXiv:2305.10924}
}

@article{salimans2022progressive,
  title        = {Progressive distillation for fast sampling of diffusion models},
  author       = {Salimans, Tim and Ho, Jonathan},
  year         = 2022,
  journal      = {arXiv preprint arXiv:2202.00512}
}

@article{selvaraju2024fora,
  title        = {FORA: Fast-Forward Caching in Diffusion Transformer Acceleration},
  author       = {Selvaraju, Pratheba and Ding, Tianyu and Chen, Tianyi and Zharkov, Ilya and Liang, Luming},
  year         = 2024,
  journal      = {arXiv preprint arXiv:2407.01425}
}

@inproceedings{ma2024deepcache,
  title        = {Deepcache: Accelerating diffusion models for free},
  author       = {Ma, Xinyin and Fang, Gongfan and Wang, Xinchao},
  year         = 2024,
  booktitle    = {Proceedings of the IEEE/CVF Conference on Computer Vision and Pattern Recognition},
  pages        = {15762--15772}
}

@article{li2023FasterDiffusion,
  title        = {Faster diffusion: Rethinking the role of unet encoder in diffusion models},
  author       = {Li, Senmao and Hu, Taihang and Khan, Fahad Shahbaz and Li, Linxuan and Yang, Shiqi and Wang, Yaxing and Cheng, Ming-Ming and Yang, Jian},
  year         = 2023,
  journal      = {arXiv preprint arXiv:2312.09608}
}

@article{chen2024delta-dit,
  title        = {$\Delta$-DiT: A Training-Free Acceleration Method Tailored for Diffusion Transformers},
  author       = {Chen, Pengtao and Shen, Mingzhu and Ye, Peng and Cao, Jianjian and Tu, Chongjun and Bouganis, Christos-Savvas and Zhao, Yiren and Chen, Tao},
  year         = 2024,
  journal      = {arXiv preprint arXiv:2406.01125}
}

@inproceedings{bolya2023tomesd,
  title        = {Token merging for fast stable diffusion},
  author       = {Bolya, Daniel and Hoffman, Judy},
  year         = 2023,
  booktitle    = {Proceedings of the IEEE/CVF conference on computer vision and pattern recognition},
  pages        = {4599--4603}
}

@inproceedings{songDDIM,
  title        = {Denoising Diffusion Implicit Models},
  author       = {Song, Jiaming and Meng, Chenlin and Ermon, Stefano},
  year         = 2021,
  booktitle    = {International Conference on Learning Representations}
}

@inproceedings{peebles2023dit,
  title        = {Scalable diffusion models with transformers},
  author       = {Peebles, William and Xie, Saining},
  year         = 2023,
  booktitle    = {Proceedings of the IEEE/CVF International Conference on Computer Vision},
  pages        = {4195--4205}
}

@inproceedings{chen2023pixartalpha,
  title        = {PixArt-$\alpha$: Fast Training of Diffusion Transformer for Photorealistic Text-to-Image Synthesis},
  author       = {Junsong Chen and Jincheng Yu and Chongjian Ge and Lewei Yao and Enze Xie and Yue Wu and Zhongdao Wang and James Kwok and Ping Luo and Huchuan Lu and Zhenguo Li},
  year         = 2024,
  booktitle    = {International Conference on Learning Representations}
}

@misc{chen2024pixartsigma,
  title        = {PixArt-$\Sigma$: Weak-to-Strong Training of Diffusion Transformer for 4K Text-to-Image Generation},
  author       = {Junsong Chen and Chongjian Ge and Enze Xie and Yue Wu and Lewei Yao and Xiaozhe Ren and Zhongdao Wang and Ping Luo and Huchuan Lu and Zhenguo Li},
  year         = 2024,
  eprint       = {2403.04692},
  archiveprefix = {arXiv},
  primaryclass = {cs.CV}
}

@software{opensora,
  title        = {Open-Sora: Democratizing Efficient Video Production for All},
  author       = {Zangwei Zheng and Xiangyu Peng and Tianji Yang and Chenhui Shen and Shenggui Li and Hongxin Liu and Yukun Zhou and Tianyi Li and Yang You},
  year         = 2024,
  month        = {March},
  url          = {https://github.com/hpcaitech/Open-Sora}
}

@article{ho2020DDPM,
  title        = {Denoising diffusion probabilistic models},
  author       = {Ho, Jonathan and Jain, Ajay and Abbeel, Pieter},
  year         = 2020,
  journal      = {Advances in neural information processing systems},
  volume       = 33,
  pages        = {6840--6851}
}

@inproceedings{sohl2015deep,
  title        = {Deep unsupervised learning using nonequilibrium thermodynamics},
  author       = {Sohl-Dickstein, Jascha and Weiss, Eric and Maheswaranathan, Niru and Ganguli, Surya},
  year         = 2015,
  booktitle    = {International conference on machine learning},
  pages        = {2256--2265},
  organization = {PMLR}
}

@inproceedings{ronneberger2015unet,
  title        = {U-net: Convolutional networks for biomedical image segmentation},
  author       = {Ronneberger, Olaf and Fischer, Philipp and Brox, Thomas},
  year         = 2015,
  booktitle    = {Medical image computing and computer-assisted intervention--MICCAI 2015: 18th international conference, Munich, Germany, October 5-9, 2015, proceedings, part III 18},
  pages        = {234--241},
  organization = {Springer}
}

@article{zou2024accelerating,
  title        = {Accelerating Diffusion Transformers with Token-wise Feature Caching},
  author       = {Zou, Chang and Liu, Xuyang and Liu, Ting and Huang, Siteng and Zhang, Linfeng},
  year         = 2024,
  journal      = {arXiv preprint arXiv:2410.05317}
}

@article{lu2022dpm,
  title        = {Dpm-solver: A fast ode solver for diffusion probabilistic model sampling in around 10 steps},
  author       = {Lu, Cheng and Zhou, Yuhao and Bao, Fan and Chen, Jianfei and Li, Chongxuan and Zhu, Jun},
  year         = 2022,
  journal      = {Advances in Neural Information Processing Systems},
  volume       = 35,
  pages        = {5775--5787}
}

@article{lu2022dpm++,
  title        = {Dpm-solver++: Fast solver for guided sampling of diffusion probabilistic models},
  author       = {Lu, Cheng and Zhou, Yuhao and Bao, Fan and Chen, Jianfei and Li, Chongxuan and Zhu, Jun},
  year         = 2022,
  journal      = {arXiv preprint arXiv:2211.01095}
}

@inproceedings{
zheng2023dpmsolvervF,
title={{DPM}-Solver-v3: Improved Diffusion {ODE} Solver with Empirical Model Statistics},
author={Kaiwen Zheng and Cheng Lu and Jianfei Chen and Jun Zhu},
booktitle={Thirty-seventh Conference on Neural Information Processing Systems},
year={2023},
url={https://openreview.net/forum?id=9fWKExmKa0}
}

@inproceedings{kim2024tofu,
  title        = {Token fusion: Bridging the gap between token pruning and token merging},
  author       = {Kim, Minchul and Gao, Shangqian and Hsu, Yen-Chang and Shen, Yilin and Jin, Hongxia},
  year         = 2024,
  booktitle    = {Proceedings of the IEEE/CVF Winter Conference on Applications of Computer Vision},
  pages        = {1383--1392}
}

@article{li2024snapfusion,
  title        = {Snapfusion: Text-to-image diffusion model on mobile devices within two seconds},
  author       = {Li, Yanyu and Wang, Huan and Jin, Qing and Hu, Ju and Chemerys, Pavlo and Fu, Yun and Wang, Yanzhi and Tulyakov, Sergey and Ren, Jian},
  year         = 2024,
  journal      = {Advances in Neural Information Processing Systems},
  volume       = 36
}

@inproceedings{shang2023post,
  title        = {Post-training quantization on diffusion models},
  author       = {Shang, Yuzhang and Yuan, Zhihang and Xie, Bin and Wu, Bingzhe and Yan, Yan},
  year         = 2023,
  booktitle    = {Proceedings of the IEEE/CVF conference on computer vision and pattern recognition},
  pages        = {1972--1981}
}

@inproceedings{song2023consistency,
  title        = {Consistency Models},
  author       = {Song, Yang and Dhariwal, Prafulla and Chen, Mark and Sutskever, Ilya},
  year         = 2023,
  booktitle    = {International Conference on Machine Learning},
  pages        = {32211--32252},
  organization = {PMLR}
}

@inproceedings{refitiedflow,
  title        = {Flow Straight and Fast: Learning to Generate and Transfer Data with Rectified Flow},
  author       = {Liu, Xingchao and Gong, Chengyue and others},
  year         = 2023,
  booktitle    = {The Eleventh International Conference on Learning Representations}
}

@article{hessel2021clipscore,
  title        = {Clipscore: A reference-free evaluation metric for image captioning},
  author       = {Hessel, Jack and Holtzman, Ari and Forbes, Maxwell and Bras, Ronan Le and Choi, Yejin},
  year         = 2021,
  journal      = {arXiv preprint arXiv:2104.08718}
}

@misc{flux2024,
  title        = {FLUX},
  author       = {Black Forest Labs},
  year         = 2024,
  howpublished = {\url{https://github.com/black-forest-labs/flux}}
}

@inproceedings{
yang2025cogvideox,
title={CogVideoX: Text-to-Video Diffusion Models with An Expert Transformer},
author={Zhuoyi Yang and Jiayan Teng and Wendi Zheng and Ming Ding and Shiyu Huang and Jiazheng Xu and Yuanming Yang and Wenyi Hong and Xiaohan Zhang and Guanyu Feng and Da Yin and Xiaotao Gu and Yuxuan.Zhang and Weihan Wang and Yean Cheng and Bin Xu and Yuxiao Dong and Jie Tang},
booktitle={The Thirteenth International Conference on Learning Representations},
year={2025},
url={https://openreview.net/forum?id=LQzN6TRFg9}
}

@INPROCEEDINGS{10377259,
  author={Li, Xiuyu and Liu, Yijiang and Lian, Long and Yang, Huanrui and Dong, Zhen and Kang, Daniel and Zhang, Shanghang and Keutzer, Kurt},
  booktitle={2023 IEEE/CVF International Conference on Computer Vision (ICCV)}, 
  title={Q-Diffusion: Quantizing Diffusion Models}, 
  year={2023},
  volume={},
  number={},
  pages={17489-17499},
  doi={10.1109/ICCV51070.2023.01608}}

@misc{zhang2024tokenpruningcachingbetter,
      title={Token Pruning for Caching Better: 9 Times Acceleration on Stable Diffusion for Free}, 
      author={Evelyn Zhang and Bang Xiao and Jiayi Tang and Qianli Ma and Chang Zou and Xuefei Ning and Xuming Hu and Linfeng Zhang},
      year={2024},
      eprint={2501.00375},
      archivePrefix={arXiv},
      primaryClass={cs.CV},
      url={https://arxiv.org/abs/2501.00375}, 
}

@misc{zou2024DuCa,
      title={Accelerating Diffusion Transformers with Dual Feature Caching}, 
      author={Chang Zou and Evelyn Zhang and Runlin Guo and Haohang Xu and Conghui He and Xuming Hu and Linfeng Zhang},
      year={2024},
      eprint={2412.18911},
      archivePrefix={arXiv},
      primaryClass={cs.LG},
      url={https://arxiv.org/abs/2412.18911}, 
}

@inproceedings{zhang2025sito,
  title={Training-Free and Hardware-Friendly Acceleration for Diffusion Models via Similarity-based Token Pruning},
  author={Zhang, Evelyn and Tang, Jiayi and Ning, Xuefei and Zhang, Linfeng},
  booktitle={Proceedings of the AAAI Conference on Artificial Intelligence},
  year={2025}
}

@misc{liu2025regionadaptivesamplingdiffusiontransformers,
      title={Region-Adaptive Sampling for Diffusion Transformers}, 
      author={Ziming Liu and Yifan Yang and Chengruidong Zhang and Yiqi Zhang and Lili Qiu and Yang You and Yuqing Yang},
      year={2025},
      eprint={2502.10389},
      archivePrefix={arXiv},
      primaryClass={cs.CV},
      url={https://arxiv.org/abs/2502.10389}, 
}

@misc{liu2024timestep,
    title={Timestep Embedding Tells: It's Time to Cache for Video Diffusion Model},
    author={Feng Liu and Shiwei Zhang and Xiaofeng Wang and Yujie Wei and Haonan Qiu and Yuzhong Zhao and Yingya Zhang and Qixiang Ye and Fang Wan},
    year={2024},
    eprint={2411.19108},
    archivePrefix={arXiv},
    primaryClass={cs.CV}
}

@misc{cheng2025catpruningclusterawaretoken,
      title={CAT Pruning: Cluster-Aware Token Pruning For Text-to-Image Diffusion Models}, 
      author={Xinle Cheng and Zhuoming Chen and Zhihao Jia},
      year={2025},
      eprint={2502.00433},
      archivePrefix={arXiv},
      primaryClass={cs.CV},
      url={https://arxiv.org/abs/2502.00433}, 
}

@misc{sun2025unicpunifiedcachingpruning,
      title={UniCP: A Unified Caching and Pruning Framework for Efficient Video Generation}, 
      author={Wenzhang Sun and Qirui Hou and Donglin Di and Jiahui Yang and Yongjia Ma and Jianxun Cui},
      year={2025},
      eprint={2502.04393},
      archivePrefix={arXiv},
      primaryClass={cs.CV},
      url={https://arxiv.org/abs/2502.04393}, 
}

@inproceedings{kim2025ditto,
  author = {Sungbin Kim and Hyunwuk Lee and Wonho Cho and Mincheol Park and Won Woo Ro},
  title = {Ditto: Accelerating Diffusion Model via Temporal Value Similarity},
  booktitle = {Proceedings of the 2025 IEEE International Symposium on High-Performance Computer Architecture (HPCA)},
  year = {2025},
  publisher = {IEEE},
}

@misc{zhu2024dipgo,
    title={DiP-GO: A Diffusion Pruner via Few-step Gradient Optimization},
    author={Haowei Zhu and Dehua Tang and Ji Liu and Mingjie Lu and Jintu Zheng and Jinzhang Peng and Dong Li and Yu Wang and Fan Jiang and Lu Tian and Spandan Tiwari and Ashish Sirasao and Jun-Hai Yong and Bin Wang and Emad Barsoum},
    year={2024},
    eprint={2410.16942},
    archivePrefix={arXiv},
    primaryClass={cs.CV}
}

@misc{VBench,
	title = {{VBench}: {Comprehensive} {Benchmark} {Suite} for {Video} {Generative} {Models}},
	shorttitle = {{VBench}},
	url = {http://arxiv.org/abs/2311.17982},
	doi = {10.48550/arXiv.2311.17982},
	urldate = {2025-02-27},
	publisher = {arXiv},
	author = {Huang, Ziqi and He, Yinan and Yu, Jiashuo and Zhang, Fan and Si, Chenyang and Jiang, Yuming and Zhang, Yuanhan and Wu, Tianxing and Jin, Qingyang and Chanpaisit, Nattapol and Wang, Yaohui and Chen, Xinyuan and Wang, Limin and Lin, Dahua and Qiao, Yu and Liu, Ziwei},
	month = nov,
	year = {2023},
	note = {arXiv:2311.17982 [cs]},
}

@misc{sun_hunyuan-large_2024,
	title = {Hunyuan-Large: An Open-Source {MoE} Model with 52 Billion Activated Parameters by Tencent},
	url = {http://arxiv.org/abs/2411.02265},
	doi = {10.48550/arXiv.2411.02265},
	shorttitle = {Hunyuan-Large},

	number = {{arXiv}:2411.02265},
	publisher = {{arXiv}},
	author = {Sun, Xingwu and Chen, Yanfeng and Huang and others},
	urldate = {2025-03-01},
	date = {2024-11-06},
	eprinttype = {arxiv},
	eprint = {2411.02265 [cs]},

}

@misc{liu2025reusingforecastingacceleratingdiffusion,
      title={From Reusing to Forecasting: Accelerating Diffusion Models with TaylorSeers}, 
      author={Jiacheng Liu and Chang Zou and Yuanhuiyi Lyu and Junjie Chen and Linfeng Zhang},
      year={2025},
      eprint={2503.06923},
      archivePrefix={arXiv},
      primaryClass={cs.CV},
      url={https://arxiv.org/abs/2503.06923}, 
}

@misc{lv2025fastercachetrainingfreevideodiffusion,
      title={FasterCache: Training-Free Video Diffusion Model Acceleration with High Quality}, 
      author={Zhengyao Lv and Chenyang Si and Junhao Song and Zhenyu Yang and Yu Qiao and Ziwei Liu and Kwan-Yee K. Wong},
      year={2025},
      eprint={2410.19355},
      archivePrefix={arXiv},
      primaryClass={cs.CV},
      url={https://arxiv.org/abs/2410.19355}, 
}

@inproceedings{saharia2022drawbench,
  title={Photorealistic Text-to-Image Diffusion Models with Deep Language Understanding},
  author={Saharia, Chitwan and Chan, William and Saxena, Saurabh and Li, Lala and Wang, Jay and Ghasemipour, Karim and Gontijo Lopes, Raphael and Lee, Burcu and Gontijo Lopes, Ekin and He, Jonathan and et al.},
  booktitle={NeurIPS},
  year={2022}
}

@article{xu2023imagereward,
  title={ImageReward: Learning and Evaluating Human Preferences for Text-to-Image Generation},
  author={Xu, Jiazheng and Li, Xiao and Xu, Guoli and Zhang, Yuan and Zhang, Xinyu and Zhou, Qishuo and Wang, Yanan and Liu, Qiming and Zhang, Yunjie and He, Yuqing and others},
  journal={arXiv preprint arXiv:2304.05977},
  year={2023}
}

@article{yuan2024ditfastattnattentioncompressiondiffusion,
  title={DiTFastAttn: Attention Compression for Diffusion Transformer Models},
  author={Yuan, Zhihang and Zhang, Hanling and Lu, Pu and Ning, Xuefei and Zhang, Linfeng and Zhao, Tianchen and Yan, Shengen and Dai, Guohao and Wang, Yu},
  journal={arXiv preprint arXiv:2406.08552},
  year={2024}
}

@article{zhao2025realtimevideogenerationpyramid,
  title={Real-Time Video Generation with Pyramid Attention Broadcast},
  author={Zhao, Xuanlei and Jin, Xiaolong and Wang, Kai and You, Yang},
  journal={arXiv preprint arXiv:2408.12588},
  year={2024}
}

@misc{zheng2025forecastcalibratefeaturecaching,
  title={Forecast then Calibrate: Feature Caching as ODE for Efficient Diffusion Transformers}, 
  author={Shikang Zheng and Liang Feng and Xinyu Wang and Qinming Zhou and Peiliang Cai and Chang Zou and Jiacheng Liu and Yuqi Lin and Junjie Chen and Yue Ma and Linfeng Zhang},
  year={2025},
  eprint={2508.16211},
  archivePrefix={arXiv},
  primaryClass={cs.CV},
  url={https://arxiv.org/abs/2508.16211}, 
}

@inproceedings{zhang2025spargeattn,
  title={Spargeattn: Accurate sparse attention accelerating any model inference},
  author={Zhang, Jintao and Xiang, Chendong and Huang, Haofeng and Wei, Jia and Xi, Haocheng and Zhu, Jun and Chen, Jianfei},
  booktitle={International Conference on Machine Learning (ICML)},
  year={2025}
}

@inproceedings{zhang2025sageattention,
  title={SageAttention: Accurate 8-Bit Attention for Plug-and-play Inference Acceleration}, 
  author={Zhang, Jintao and Wei, Jia and Zhang, Pengle and Zhu, Jun and Chen, Jianfei},
  booktitle={International Conference on Learning Representations (ICLR)},
  year={2025}
}

@inproceedings{zhang2024sageattention2,
  title={Sageattention2: Efficient attention with thorough outlier smoothing and per-thread int4 quantization},
  author={Zhang, Jintao and Huang, Haofeng and Zhang, Pengle and Wei, Jia and Zhu, Jun and Chen, Jianfei},
  booktitle={International Conference on Machine Learning (ICML)},
  year={2025}
}
}


\end{document}